\documentclass[conference]{IEEEtran}
\IEEEoverridecommandlockouts
\pdfoutput=1
\usepackage{cite}
\usepackage{bm} 
\usepackage{color}
\usepackage{amsmath}
\usepackage{booktabs}
\usepackage{adjustbox}
\usepackage{gensymb}
\usepackage{amssymb}
\usepackage{float}
\usepackage{lineno,hyperref}
\usepackage{graphicx}
\usepackage{color}
\usepackage{tabu}
\usepackage{multirow}
\usepackage{flushend}
\graphicspath{{./figures/}}
\DeclareGraphicsExtensions{.pdf,.jpeg,.png,.jpg,.tif}
\usepackage[caption=false]{subfig}
\makeatletter
\def\endthebibliography{%
  \def\@noitemerr{\@latex@warning{Empty `thebibliography' environment}}%
  \endlist
}
\makeatother

\begin{document}

\title{Multi-fidelity power flow solver}

\author{
    \IEEEauthorblockN{Sam Yang\IEEEauthorrefmark{1}\thanks{This work was supported through the INL Laboratory Directed Research \& Development (LDRD) Program under DOE Idaho Operations Office Contract DE-AC07-05ID14517. This research made use of Idaho National Laboratory computing resources which are supported by the Office of Nuclear Energy of the U.S. Department of Energy and the Nuclear Science User Facilities under Contract No. DE-AC07-05ID14517.}, Bjorn Vaagensmith\IEEEauthorrefmark{3}, Deepika Patra\IEEEauthorrefmark{3}, Ryan Hruska\IEEEauthorrefmark{3}, Tyler Phillips\IEEEauthorrefmark{3}}
    \IEEEauthorblockA{\IEEEauthorrefmark{1}Center for Advanced Power Systems, Florida State University, Tallahassee, Florida 32310}
    \IEEEauthorblockA{\IEEEauthorrefmark{3}Idaho National Laboratory, Idaho Falls, Idaho 83415}
    \text{Email}: syang@caps.fsu.edu, \{bjorn.vaagensmith, ryan.hruska, tyler.phillips\}@inl.gov
}
\maketitle


\begin{abstract}
We propose a multi-fidelity neural network (MFNN) tailored for rapid high-dimensional grid power flow simulations and contingency analysis with scarce high-fidelity contingency data. The proposed model comprises two networks---the first one trained on DC approximation as low-fidelity data and coupled to a high-fidelity neural net trained on both low- and high-fidelity power flow data. Each network features a latent module which parametrizes the model by a discrete grid topology vector for generalization (e.g., $n$ power lines with $k$ disconnections or contingencies, if any), and the targeted high-fidelity output is a weighted sum of linear and nonlinear functions. We tested the model on 14- and 118-bus test cases and evaluated its performance based on the $n-k$ power flow prediction accuracy with respect to imbalanced contingency data and high-to-low-fidelity sample ratio. The results presented herein demonstrate MFNN's potential and its limits with up to two orders of magnitude faster and more accurate power flow solutions than DC approximation.
\end{abstract}


\IEEEpeerreviewmaketitle

\section{Introduction}
\IEEEPARstart{T}{he} quest for resilient power grids has driven utilities to consider all conceivable vulnerabilities and comply with strict standards such as NERC TPL-001. Transmission systems, for example, are required to withstand any single component failure (i.e., any $n-1$ contingency) by providing appropriate remedial actions without a major interruption. Contingency analysis and integrated resource planing (IRP) have thus become a critical step in grid design and maintenance, and power systems engineers presently rely on a selective number of power flow solvers, namely the Newton-Raphson (NR) method, DC approximation, and fast decoupled method, to discover vulnerabilities subject to grid component failures. 

Numerical solvers for power flow simulations must warrant numerical stability, accuracy, and computational efficiency, but their competing nature has forced engineers and utilities to search for an optimal balance between their trade-offs. The NR method, for instance, considers both the real and reactive components of power flow equations, and its solutions are typically regarded as the ground truth in this domain. The fast decoupled method is computationally cheaper than the NR method but suffers from convergence issues \cite{Wang2000DPF}. Commercial and open-source power simulation tools such as PowerWorld, MATPOWER \cite{zimmerman2010matpower}, and Pandapower \cite{thurner2018pandapower} leverage the DC approximation to initialize more accurate nonlinear solvers for specific problems. Present utilities exploit these conventional tools to perform power flow simulations, $n-k$ contingency analysis, and IRP.

The growing complexity and push toward distributed, renewable, and automated power grid operation \cite{PROSTEJOVSKY2019105883} with minimal maintenance and human intervention are promoting the need for high-order contingency analysis. While such assessments are reasonable for smaller grids, the high computational cost associated with traditional solvers often prohibit $n-2$ and greater contingency analysis of large grids. For example, PowerWorld sweeps through all $n-1$ contingencies on Texas 2000-bus system in under 2 min (on an Intel(R) Xenon(R) E-2286M and 128 GB of 2667 MHz DDR4) while it takes more than three days to obtain all $n-2$ contingencies. Novel algorithms have therefore been proposed to overcome this challenge \cite{molzahn2019survey, yang2020MLreview}, from which machine learning (ML)-based approaches, particularly deep learning, have emerged as prominent alternatives \cite{rudin2011machine, sabri2015improvement, donnot2017introducing, donnot2019leap, donon2020neural, jeddi2021physics, kody2021modeling, falconer2021leveraging}.

Donnot et al. \cite{donnot2019leap} introduced the Latent Encoding of Atypical Perturbations (LEAP) net which embeds grid topology and its changes for generalization. Inspired by ResNET, the LEAP net was able to learn topological changes and infer power flow solutions accordingly based on inductive bias. Donon et. al. \cite{donon2020neural} proposed a physics-informed graph neural net (PIGNN) which computed power flow by directly minimizing the violation of Kirchhoff’s law at each bus during training. The PIGNN was able to solve for power flow more accurately than DC approximation with $0.01$--$10\%$ error deviations from the NR method and was robust to variations of injections, power grid topology, and line characteristics. Ref.~\cite{donon2020neural} also demonstrated PIGNN's transfer learning capabilities for estimating power flow in grids it was never trained for. 

Jeddi and Shafieezadeh \cite{jeddi2021physics} also introduced a graph attention-based PIGNN and evaluated its performance on the IEEE 9-, 14-, 30-, and 118-bus test cases. The proposed model was about an order of magnitude more accurate and computationally faster than a regular GNN. Falconer and Mones \cite{falconer2021leveraging} evaluated fully-connected NNs (FNNs), convolutional NNs (CNNs), and GNNs for predicting optimal generator set-points and active set of constraints. The authors demonstrated marginal utility of employing CNN and GNN compared to FNN for a fixed grid topology, whereas GNNs were able to account for the topological changes (e.g., modeling transmission line contingency) and outperform both FNN and CNN.

According to our literature review, deep learning-based power flow models have evolved extensively in recent years especially with the introduction of PIGNNs. PIGNNs have enabled the creation of physically-constrained and explainable GNNs while bridging the gap between classical physics-based power flow models and purely data-driven models. While PIGNNs are gaining more traction for power flow simulations, here we provide an alternative solution by exploiting readily available low-fidelity power flow data (e.g., DC approximation) to provide solutions comparable to the NR method with scarce high-fidelity contingency data. 

We propose herein a multi-fidelity neural network (MFNN) tailored for rapid high-dimensional power flow simulations with minimal high-fidelity training data and straightforward embedding of grid topology. The objective of this work is to explore the advantages and limits of MFNNs for power flow analysis which have not yet been done to the best of our knowledge. We tested the model on 14- and 118-bus test cases and evaluated its performance based on the $n-k$ power flow prediction accuracy with respect to imbalanced contingency data and high-to-low-fidelity sample ratio. The remainder of this paper is organized as follows: Section~\ref{sec:mfleap} describes the MFNN followed by a test problem setup and results in Section \ref{sec:test_problem} and \ref{sec:results}, respectively. We then summarize our findings in Section~\ref{sec:conclusion}.

\section{Multi-Fidelity Power Flow Solver\label{sec:mfleap}}
\begin{figure}[h!]
    \centering
    \includegraphics[width=1\hsize]{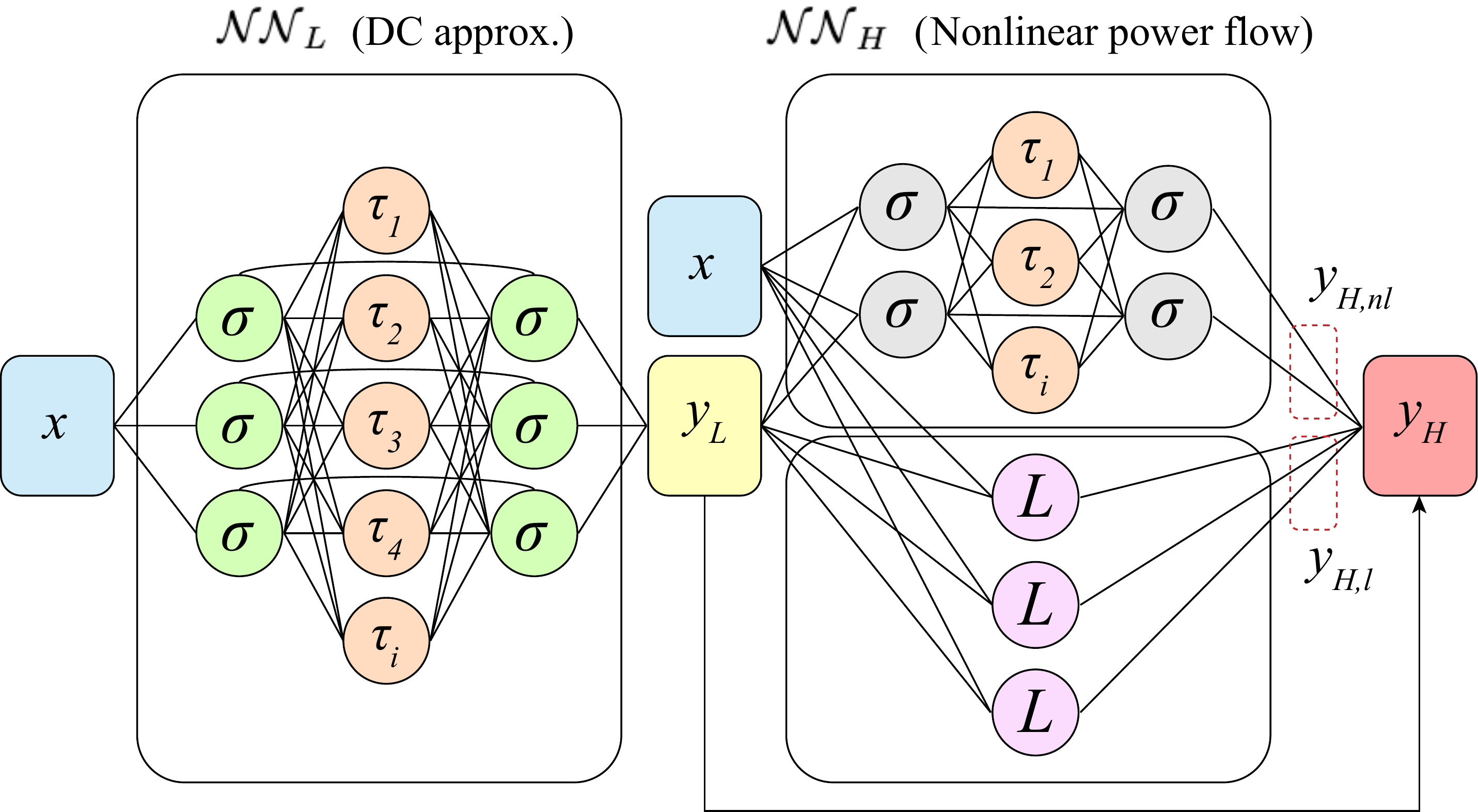}
    \caption{Multi-fidelity neural network architecture where $L$ and $\sigma$ are linear and nonlinear activation functions while $\tau$ and the subscripts $L$ and $H$ denote power line service status, low, and high fidelity, respectively.}
    \label{fig:mfnn}
\end{figure}

The proposed MFNN illustrated in Fig.~\ref{fig:mfnn} combines the residual MFNN \cite{lu2020extraction} and the LEAP net \cite{donnot2019leap}. The model comprises two $\mathcal{NN}$s---the first one trained on DC approximation (low-fidelity data) and coupled to a high-fidelity $\mathcal{NN}$ trained on both low and high-fidelity power flow data. Each $\mathcal{NN}$ features a latent module $\bm{\tau}$ which parametrizes the model by a discrete grid topology vector for generalization (e.g., $n$ power lines with $k$ disconnections or contingencies, if any), and the targeted high-fidelity output is a weighted sum of linear and nonlinear functions. $\bm{\tau}=\lbrace \tau_1,\tau_2,\ldots,\tau_i\rbrace$ in Fig.~\ref{fig:mfnn} represents line service status, e.g., $\tau_i=0$ (out of service); $\tau_i=1$ (in service).

The LEAP net exhibits \textit{super-generalization} properties which is desired for high-order contingency analysis subject to additive and multiplicative perturbations. Ref.~\cite{donon2020leap} defines \textit{super-generalization} as the capability of a neural net $y=\mathcal{NN}\left(\bm{x},\bm{\tau}\right)$ to approximate $y^*=S(\bm{x},\bm{\tau})$ when the natural observational distribution of $\bm{\tau}$ is very imbalanced, with a peak around a reference topology $\bm{\tau}_0$. The \textit{super-generalization} capability of LEAP net is formally proved in Ref.~\cite{donon2020leap} and is described as
\begin{equation}
    y_i = \mathbf{D}(\mathbf{E}(\bm{x})) + 
    \mathbf{d}(\mathbf{e}(\mathbf{E}(\bm{x})\odot\bm{\tau})) \in\mathbb{R},
\end{equation}
where $\mathbf{E}$, $\mathbf{e}$, $\mathbf{D}$, and $\mathbf{d}$ are differentiable functions. The LEAP net, unlike similar $\mathcal{NN}$ architectures such as the ResNET, implicitly encodes topological changes in the grid by inferring the relationship between serviced lines, generators, and loads from $\bm{\tau}$.

The high-fidelity output $y_H$ in Fig.~\ref{fig:mfnn} is a weighted sum of a linear function and a nonlinear function as \cite{lu2020extraction} 
\begin{equation}
\begin{split}
     y_H=\alpha_L y_L+
     \varepsilon(\tanh{\alpha_1}\cdot y_{H,l}\left(\bm{x},y_L\right)+\\
     \tanh{\alpha_2}\cdot y_{H,nl}\left(\bm{x},y_L\right)).
\end{split}
\end{equation}
The coefficients $\alpha_1$ and $\alpha_2$ are trained parameters along with $\alpha_L$ which is the ratio of the high-to-low-fidelity outputs. Here $\varepsilon=0.1$ since $y_L$ and $y_H$ are of the same magnitude. The $\mathcal{NN}$s are trained by minimizing the mean squared error ($\mathrm{MSE}$) as
\begin{equation}
\begin{split}
\mathrm{MSE}=
\frac{1}{n_{y_L}}\sum_{i=1}^{n_{y_L}}\left(|y_L^{*}-y_L|^2\right)+\\
\frac{1}{n_{y_H}}\sum_{i=1}^{n_{y_H}}\left(|y_H^{*}-y_H|^2\right)+
\lambda\sum\beta_i^2 .
\end{split}
\label{eq:loss}
\end{equation}
Here $y^{*}$ denotes training data and $\lambda$ is the $\mathrm{L}_2$ regularization rates for $\beta$ which is adopted to avoid overfitting. Eq.~\eqref{eq:loss} is minimized using the Adam optimizer together with Glorot uniform and $\tanh$ as the initializer and activation function in $\mathcal{NN}_L$ and $\mathcal{NN}_{H,nl}$. As noted in Fig.~\ref{fig:mfnn}, $\mathcal{NN}_{H,l}$ has no activation function as it is used to approximate the linear component. 

The proposed model accuracy depends heavily on the training data distribution and high-to-low-fidelity sample ratio $0\leq\omega\leq 1$. In particular, the distribution of $\bm{\tau}$ and $\bm{x}$ dictates the transfer learning performance as they provide information about how much and in which direction the system actually deviates from the reference state. Similarly, higher $\omega$ will yield more accurate predictions in exchange for an increased computational cost and vice-versa. We will thus explore and quantify the effect of $\omega$ on the prediction accuracy for the test problem. 

\section{Test Problem\label{sec:test_problem}}
We tested the MFNN on the IEEE 14 bus test case which represents a portion of the Midwestern US electric power system (as of February, 1962). The test grid comprises 5 generators, 15 lines, and 11 loads. Both the low- and high-fidelity training data (i.e., DC approximation and NR solutions) were obtained using Pandapower \cite{thurner2018pandapower}, an open-source power flow solver in Python.

\begin{figure}[!h]
    \centering
    \includegraphics[width=1\hsize]{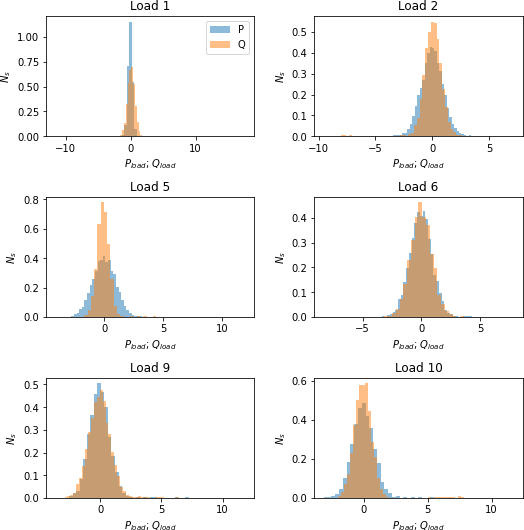}
    \caption{Training data input example--scaled real and reactive load power (not shown are $\bm{\tau}$, active power and voltage of generators).}
    \label{fig:input}
\end{figure}

\begin{figure}[!h]
    \centering
    \subfloat[14-bus]{
    \includegraphics[width=.7\hsize]{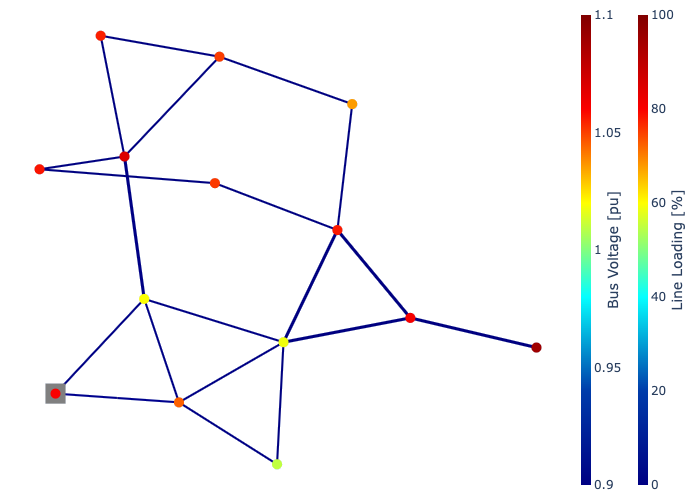}\label{fig:14_ref}}
    
    \subfloat[118-bus]{
    \includegraphics[width=.7\hsize]{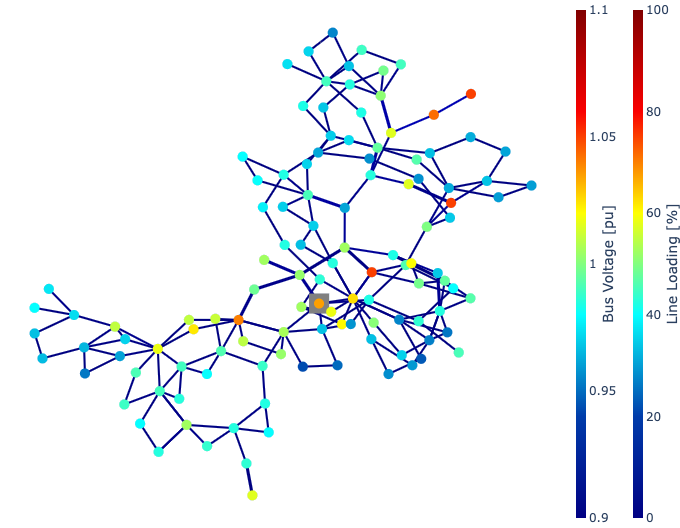}\label{fig:118_ref}}
    \caption{Bus voltage and line loading distribution with zero contingency.}
\end{figure}

The test problem consists of training the MFNN on $n-k$ contingency data for $k=1$ and $2$, normally distributed generator power $p_g$, generator voltage $v_g$, real and reactive load power $p_l$, $q_l$ (see Fig.~\ref{fig:input}), and $k$ line failure probabilities $\rho$. Here $\rho=1$ implies $k$ lines are guaranteed to fail in every simulation while opposite is the case with $\rho=0$, i.e., $\rho$ defines the contingency data imbalance. Fig.~\ref{fig:14_ref} shows the bus voltage and line loading across the 14-bus network which serves as our reference case with zero contingency. 

We trained the MFNN with $0.01\leq \omega\leq 0.9$ on a Lambda Hyperplane deep learning distributed memory system housing 4 NVIDIA A100 Tensor Core GPUs. The trained model was then used to predict power flow comprising active line power flow $p_{li}$, line current $i_{li}$, line voltage $v_{li}$, and line loading $\theta_{li}$ while evaluating its accuracy with respect to $\rho$ and $\omega$. In reference to Fig.~\ref{fig:mfnn}, our $\bm{x}=\lbrace\bm{p_g},\bm{v_g},\bm{p_l},\bm{Q_l},\bm{\tau}\rbrace$ and $y=\lbrace \bm{p}_{li}, \bm{i}_{li}, \bm{v}_{li}, \bm{\theta}_{li}\rbrace$. In addition to the 14-bus grid, we also tested the model on the IEEE 118-bus test case \cite{christie1993ieee} comprising 53 generators, 173 lines, and 99 loads as illustrated in Fig.~\ref{fig:118_ref}.

\section{Results and Discussion\label{sec:results}}
\begin{figure}[!h]
    \centering
    \includegraphics[width=.9\hsize]{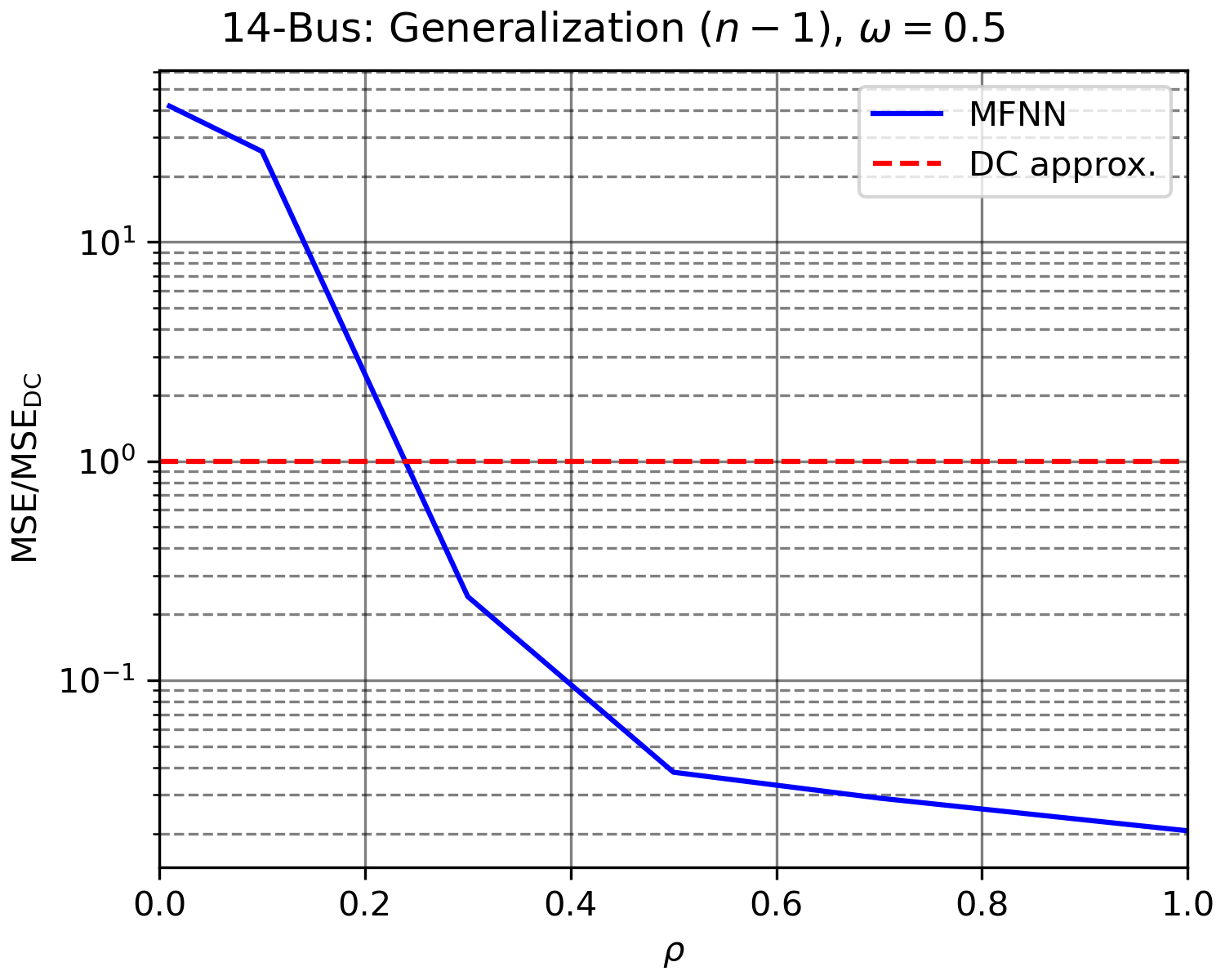}
    \caption{Effect of $\rho$ on the MFNN prediction accuracy.}
        \label{fig:result_general}
\end{figure}

The tested MFNN architecture was heuristically defined for the 14-bus grid with 8 hidden layers and about 1.2k hidden units in each $\mathcal{NN}$. Fig.~\ref{fig:result_general} shows the effect of $\rho$ on the predicted 14-bus power flow accuracy under $n-1$ contingency. With $\omega=0.5$, the MFNN began to outperform the DC approximation with only 30\% of the total training data pertaining to $n-1$ contingency. The predicted solution error becomes an order of magnitude smaller than the DC approximation as we further increase $\rho$. This is an important property and unique advantage of the proposed MFNN stemming from the LEAP net as utilities may not have enough access to neither low- nor high-fidelity data to train and fine-tune their models.

\begin{figure}[!h]
    \centering
    \subfloat[14-bus]{
    \includegraphics[width=.9\hsize]{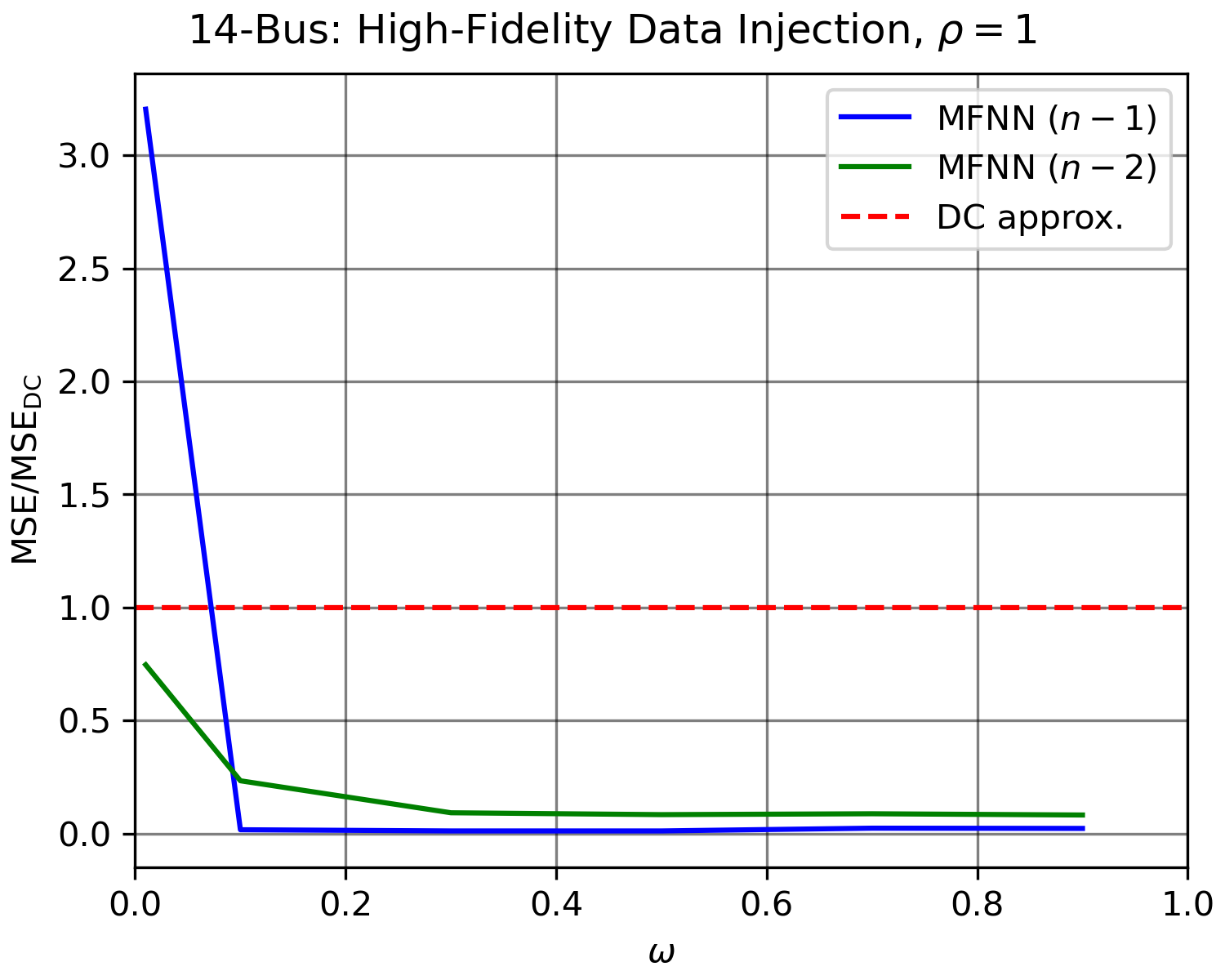}\label{fig:result_14inject}}
    
    \subfloat[118-bus]{
    \includegraphics[width=.9\hsize]{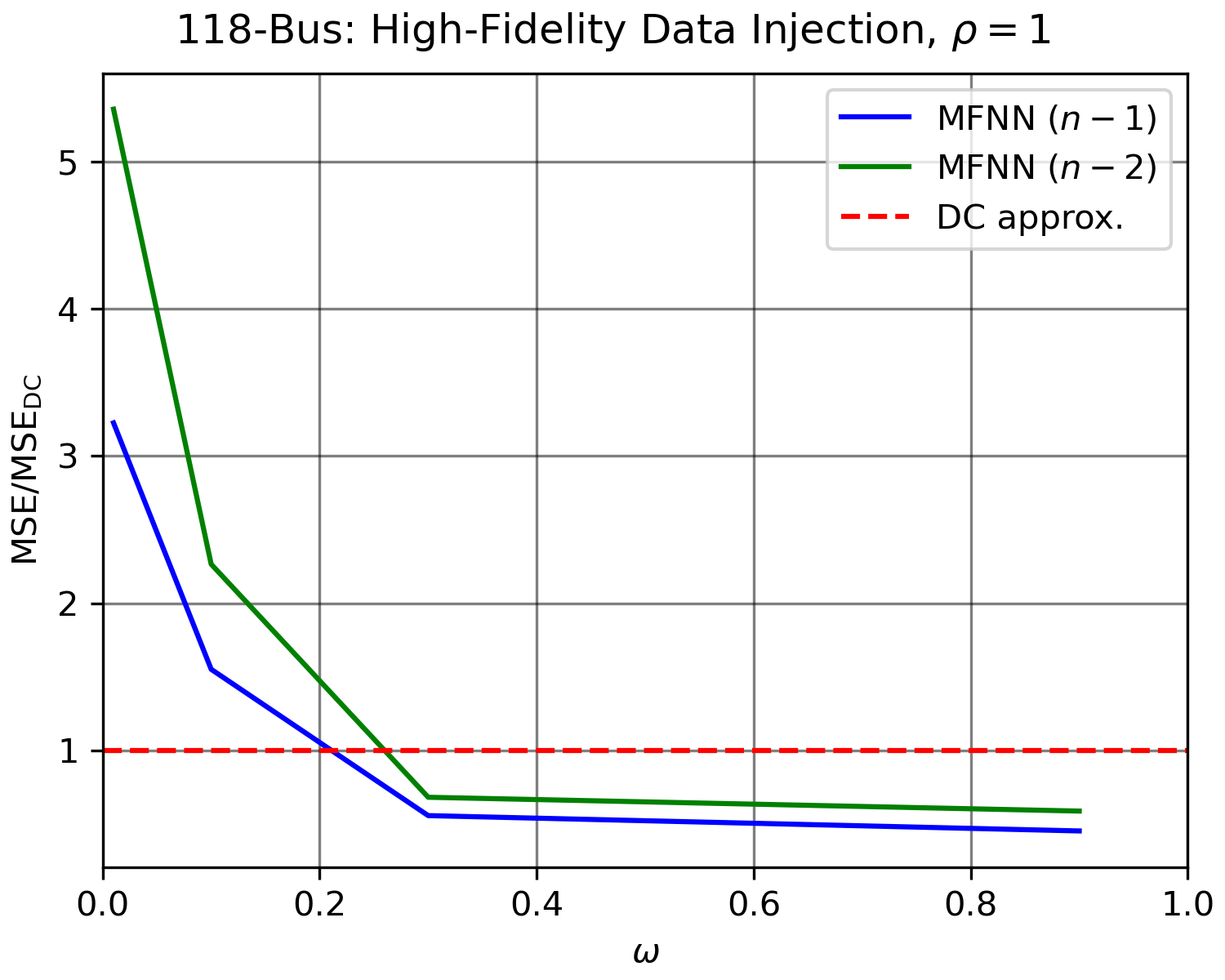}\label{fig:result_118inject}}
    \caption{Effect of $\omega$ on the model prediction accuracy.}
    \label{fig:result_omega}
\end{figure}

Now the question is, ``What happens if we have plenty of low-fidelity data while high-fidelity samples are scarce or too expensive?" The answer to this question arising in real power grid modeling and simulations is given in Fig.~\ref{fig:result_omega}, according to which the MFNN outperforms DC approximation even with a substantially small number of high-fidelity samples. In the case of 14-bus grid $\omega$ can be as small as $0.1$ to yield an order of magnitude more accurate solution than DC approximation, whereas $\omega\geq 0.3$ for the 118-bus case in Fig.~\ref{fig:result_118inject}. Fig.~\ref{fig:result_omega} essentially shows how MFNN performance degrades as the grid size and contingency order increase. This is one of the primary drawbacks of the MFNN---it does not have physical knowledge about the grid structure and constraints---which limits its accuracy and generality to specific network architecture and problem it was trained for.

\begin{figure}[!h]
    \centering
    \subfloat[14-bus]{
   \includegraphics[width=1\hsize]{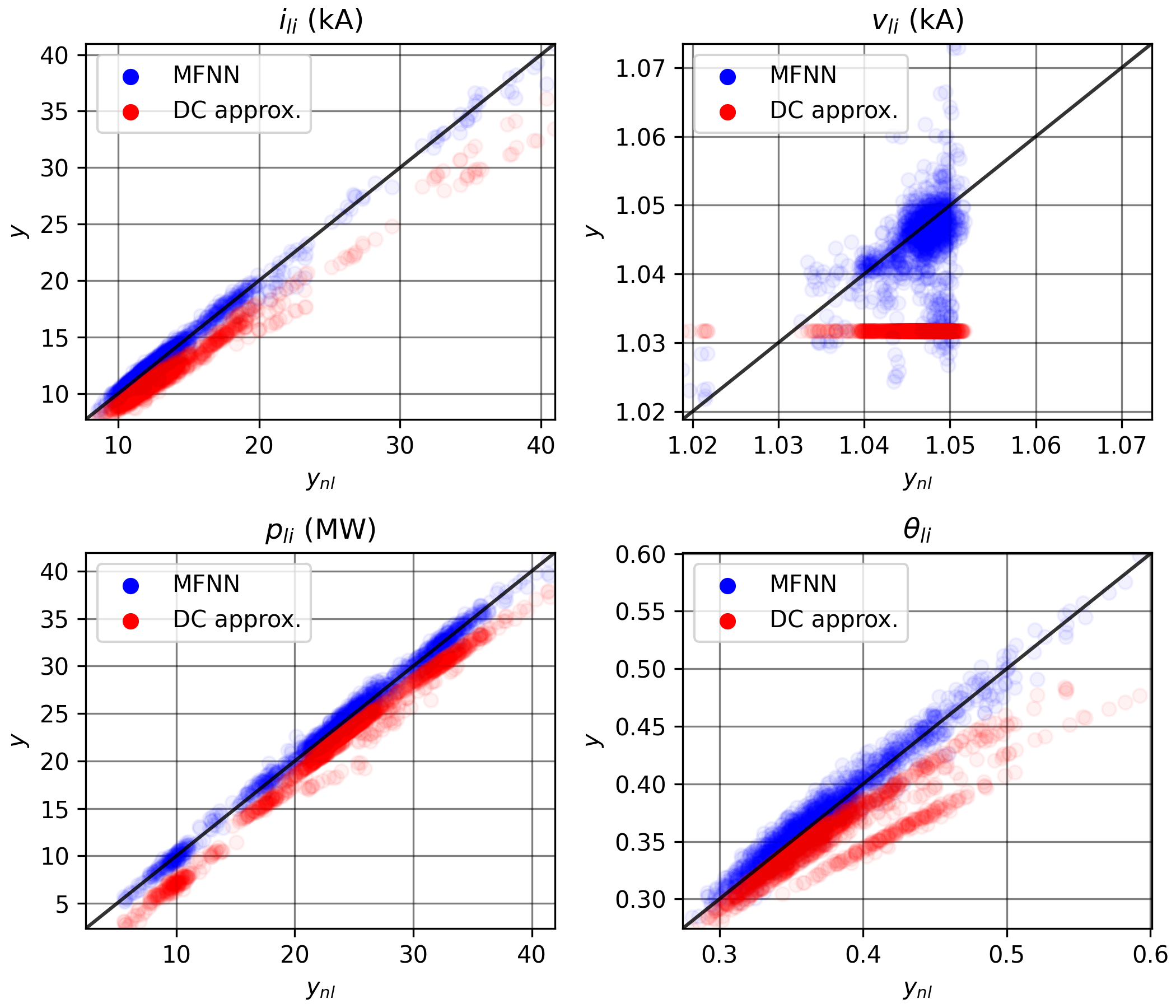}\label{fig:result_14bus}}
    
    \subfloat[118-bus]{
   \includegraphics[width=1\hsize]{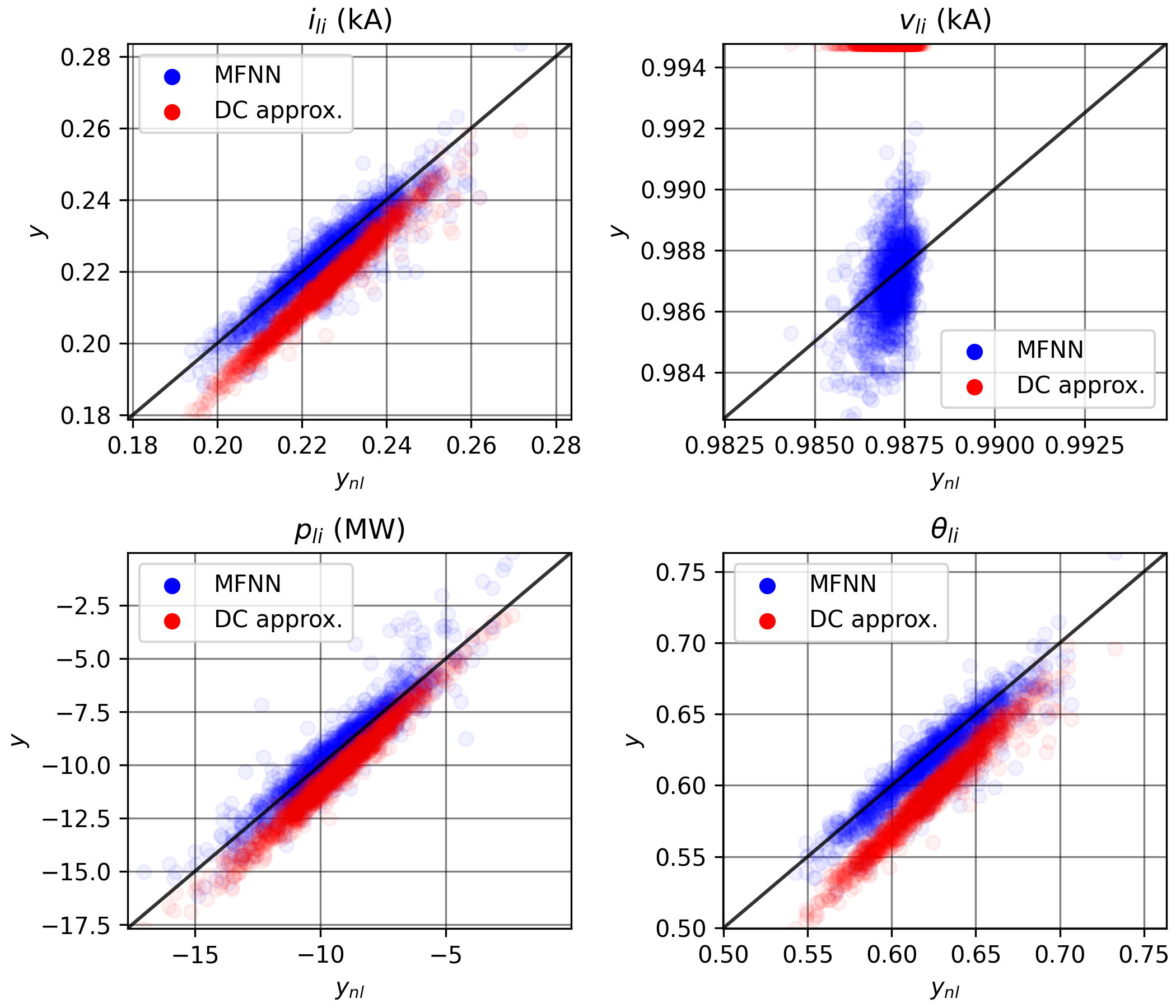}\label{fig:result_118bus}}
    \caption{Predicted line characteristics under out-of-sample $n-2$ contingencies with the MFNN trained on 118-bus, $n-2$, $\rho=1$, and $\omega=0.3$.}
    \label{fig:results}
\end{figure}

Fig.~\ref{fig:results} displays the qualitative agreement between the MFNN, DC approximation, and the NR method for the out-of-sample $n-2$ contingencies with $\rho=1$ and $\omega=0.3$. The power flow obtained by the MFNN closely follows the NR solutions for the 14-bus case (see Fig.~\ref{fig:result_14bus}), except for the line voltage ${v}_{li}$ whose maximum deviation is $\approx 2\%$ and thus deemed negligible. Such a discrepancy is also observed with the 118-bus case in Fig.~\ref{fig:result_118bus}, where $v_{li}$ of both methods are uncorrelated to the NR solutions. 

The deviation of the two methods is slightly larger for the 118-bus case with more outliers (see Fig.~\ref{fig:result_118bus}). The increased prediction error between the MFNN and the NR method in the 118-bus test case is due to the generalization error with nearly 700 target variables to predict (note that the MFNN was designed for the 14-bus case). This discrepancy can be reduced via adoption of a larger MFNN in exchange for an increased computational cost.

Table~\ref{tab:MSE_compare} compares DC approximation and MFNN predictions based on their average MSE. It is evident that the proposed MFNN outperforms DC approximation in all considered test cases. One of our pending tasks entails comparing the MFNN against other ML-based methods such as FNN, LEAP net, and PIGNN. It is also worth noting the exclusion of reactive power flow $\bm{q}_{li}$ (in $\mathrm{MVar}$) from the target output $y$ to enforce the same target variable space for low and high-fidelity data sources (recall that DC approximation do not consider reactive components). However, this must be addressed since both real and reactive power flow are critical for reliable contingency analysis and IRP. We therefore propose the implementation of a variable output mapping scheme as future work to incorporate additional information available through specific data sources as with the input mapping \cite{hebbal2021multi}. 

\renewcommand{\arraystretch}{1.2}
\begin{table}[!h]
\centering
\caption{Average MSE between true and estimated (DC and MFNN) power flow on the 14 and 118-bus test cases}
\label{tab:MSE_compare}
\begin{tabular}{lllll}
\hline
& \multicolumn{2}{c}{\textbf{14-bus}} & \multicolumn{2}{c}{\textbf{118-bus}} \\ \hline
 & \multicolumn{1}{c}{DC} & \multicolumn{1}{c}{MFNN} & \multicolumn{1}{c}{DC} & \multicolumn{1}{c}{MFNN} \\ \hline
${i}_{li}$ & 3.66 & 0.15 & $10^{-4}$ & $10^{-5}$ \\
${v}_{li}$ & $10^{-4}$ & $10^{-5}$ & $10^{-4}$ & $10^{-6}$ \\
${p}_{li}$ & 4.38 & 0.25 & 1.12 & 0.51 \\
${\theta}_{li}$ & $10^{-3}$ & $10^{-4}$ & $10^{-3}$ & $10^{-4}$ \\
\hline
\end{tabular}
\end{table}

\begin{figure}[!h]
    \centering
    \includegraphics[width=.9\hsize]{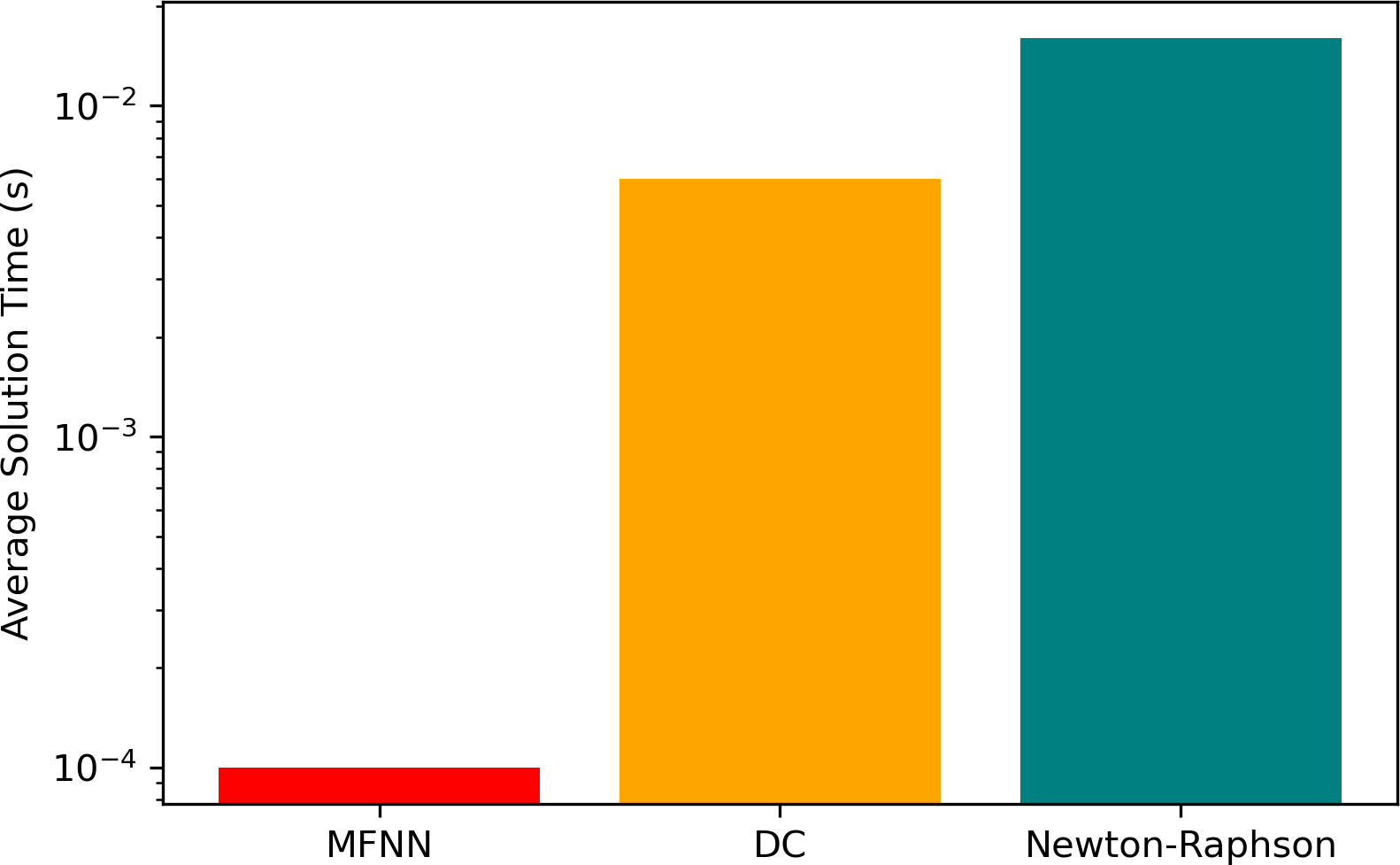}
    \caption{Average load flow time from the 118-bus $n-2$ contingency analysis.}
    \label{fig:solution_time}
\end{figure}

Fig.~\ref{fig:solution_time} compares the average computational time required by the MFNN, DC approximation, and the NR method to perform one load flow for the 118-bus $n-2$ contingency case. The MFNN outperforms other methods by nearly two orders of magnitude, which is comparable to that of PIGNNs \cite{jeddi2021physics}. Hence the MFNN is deemed suitable for real-time grid analysis and decision making in addition to its utility for high-order contingency analysis.

\section{Conclusion\label{sec:conclusion}}
Here we proposed an MFNN and explored its potential and the limits for power flow simulations and high-order contingency analysis with scarce high-fidelity contingency data. Even without any prior knowledge about the grid layout, governing equations, and physical constrains as with PIGNNs, the proposed model was able to infer power flow subject to $n-k$ contingencies up to two orders of magnitude more accurately and faster than DC approximation constituting a large fraction of the training data. However, the MFNN lacked generality in dealing with unseen grid topologies and its performance degraded quickly with respect to grid size and contingency order.

More work remains to be done for a wider adoption of the power flow MFNN. We plan to improve the MFNN by incorporating a variable output scheme for reactive line power and optimizing hyperparameters for scalability. The ultimate goal of this ongoing effort is to combine the multi-fidelity modeling approach described herein with PIGNNs \cite{jeddi2021physics} and Bayesian nets \cite{neal2012bayesian} for faster convergence, improved scalability, uncertainty quantification, and confidence-based estimation. We expect the final product to leverage the best of both data-driven and physics-based power flow modeling and simulation, which will also facilitate its application to other physics domains including fluid and gas flow in pipelines. It is possible that by applying such models to other infrastructure systems, one can achieve enhanced system operations and regional resilience by potentially avoiding disruptions that cascaded from system to system.  

A fundamental extension of this work is the prediction of both AC and DC optimal power flows (OPF) as in Ref.~\cite{falconer2021leveraging}, including security constrained OPF, which seek to provide the least-cost solutions while minimizing system constraints. The OPF problem has been shown to be non-convex and NP-Hard, which has limited its use \cite{gan2014optimal}. Efficient OPF solutions are not only essential for real-time applications, but also for optimal transmission and IRP. As utilities transition away from centralized large-scale controllable fossil fuel generators to a fleet of smaller non-emitting distributed energy resources, traditional methods used to plan future reliable and resilient delivery of electricity are potentially insufficient. 

The IRP process already requires solving a large-scale nonlinear optimization problem which often result in simplifying assumptions and the avoidance of contingency analysis altogether. Policy makers, regulators and utilities require tools that leverage computationally efficient approaches to better evaluate the complex behaviors of different portfolios that make up the bulk electric system, as well as understand their potential vulnerabilities to all threats and hazards.

\bibliographystyle{IEEEtran}
\bibliography{references}

\begin{thebibliography}{10}
\providecommand{\url}[1]{#1}
\csname url@samestyle\endcsname
\providecommand{\newblock}{\relax}
\providecommand{\bibinfo}[2]{#2}
\providecommand{\BIBentrySTDinterwordspacing}{\spaceskip=0pt\relax}
\providecommand{\BIBentryALTinterwordstretchfactor}{4}
\providecommand{\BIBentryALTinterwordspacing}{\spaceskip=\fontdimen2\font plus
\BIBentryALTinterwordstretchfactor\fontdimen3\font minus
  \fontdimen4\font\relax}
\providecommand{\BIBforeignlanguage}[2]{{%
\expandafter\ifx\csname l@#1\endcsname\relax
\typeout{** WARNING: IEEEtran.bst: No hyphenation pattern has been}%
\typeout{** loaded for the language `#1'. Using the pattern for}%
\typeout{** the default language instead.}%
\else
\language=\csname l@#1\endcsname
\fi
#2}}
\providecommand{\BIBdecl}{\relax}
\BIBdecl

\bibitem{Wang2000DPF}
L.~Wang and X.~Lin, ``Robust fast decoupled power flow,'' \emph{IEEE
  Transactions on Power Systems}, vol.~15, no.~1, pp. 208--215, 2000.

\bibitem{zimmerman2010matpower}
R.~D. Zimmerman, C.~E. Murillo-S{\'a}nchez, and R.~J. Thomas, ``Matpower:
  Steady-state operations, planning, and analysis tools for power systems
  research and education,'' \emph{IEEE Transactions on power systems}, vol.~26,
  no.~1, pp. 12--19, 2010.

\bibitem{thurner2018pandapower}
L.~Thurner, A.~Scheidler, F.~Sch{\"a}fer, J.-H. Menke, J.~Dollichon, F.~Meier,
  S.~Meinecke, and M.~Braun, ``pandapower—an open-source python tool for
  convenient modeling, analysis, and optimization of electric power systems,''
  \emph{IEEE Transactions on Power Systems}, vol.~33, no.~6, pp. 6510--6521,
  2018.

\bibitem{PROSTEJOVSKY2019105883}
\BIBentryALTinterwordspacing
A.~M. Prostejovsky, C.~Brosinsky, K.~Heussen, D.~Westermann, J.~Kreusel, and
  M.~Marinelli, ``The future role of human operators in highly automated
  electric power systems,'' \emph{Electric Power Systems Research}, vol. 175,
  p. 105883, 2019. [Online]. Available:
  \url{https://www.sciencedirect.com/science/article/pii/S0378779619302020}
\BIBentrySTDinterwordspacing

\bibitem{molzahn2019survey}
D.~K. Molzahn, I.~A. Hiskens \emph{et~al.}, ``A survey of relaxations and
  approximations of the power flow equations,'' \emph{Foundations and Trends in
  Electric Energy Systems}, vol.~4, no. 1-2, 2019.

\bibitem{yang2020MLreview}
S.~Yang, B.~Vaagensmith, and D.~Patra, ``Power grid contingency analysis with
  machine learning: A brief survey and prospects,'' in \emph{2020 Resilience
  Week (RWS)}, 2020, pp. 119--125.

\bibitem{rudin2011machine}
C.~Rudin, D.~Waltz, R.~N. Anderson, A.~Boulanger, A.~Salleb-Aouissi, M.~Chow,
  H.~Dutta, P.~N. Gross, B.~Huang, S.~Ierome \emph{et~al.}, ``Machine learning
  for the new york city power grid,'' \emph{IEEE transactions on pattern
  analysis and machine intelligence}, vol.~34, no.~2, pp. 328--345, 2011.

\bibitem{sabri2015improvement}
M.~Sabri and R.~Rezaeipour, ``Improvement estimation power flow using bayesian
  neural network,'' \emph{International Journal of Information Technology and
  Electrical Engineering}, vol.~4, pp. 32--40, 2015.

\bibitem{donnot2017introducing}
B.~Donnot, I.~Guyon, M.~Schoenauer, P.~Panciatici, and A.~Marot, ``Introducing
  machine learning for power system operation support,'' \emph{arXiv preprint
  arXiv:1709.09527}, 2017.

\bibitem{donnot2019leap}
B.~Donnot, B.~Donon, I.~Guyon, Z.~Liu, A.~Marot, P.~Panciatici, and
  M.~Schoenauer, ``Leap nets for power grid perturbations,'' \emph{arXiv
  preprint arXiv:1908.08314}, 2019.

\bibitem{donon2020neural}
B.~Donon, R.~Cl{\'e}ment, B.~Donnot, A.~Marot, I.~Guyon, and M.~Schoenauer,
  ``Neural networks for power flow: Graph neural solver,'' \emph{Electric Power
  Systems Research}, vol. 189, p. 106547, 2020.

\bibitem{jeddi2021physics}
A.~B. Jeddi and A.~Shafieezadeh, ``A physics-informed graph attention-based
  approach for power flow analysis,'' in \emph{2021 20th IEEE International
  Conference on Machine Learning and Applications (ICMLA)}.\hskip 1em plus
  0.5em minus 0.4em\relax IEEE, 2021, pp. 1634--1640.

\bibitem{kody2021modeling}
A.~Kody, S.~Chevalier, S.~Chatzivasileiadis, and D.~Molzahn, ``Modeling the ac
  power flow equations with optimally compact neural networks: Application to
  unit commitment,'' \emph{arXiv preprint arXiv:2110.11269}, 2021.

\bibitem{falconer2021leveraging}
T.~Falconer and L.~Mones, ``Leveraging power grid topology in machine learning
  assisted optimal power flow,'' \emph{arXiv preprint arXiv:2110.00306}, 2021.

\bibitem{lu2020extraction}
L.~Lu, M.~Dao, P.~Kumar, U.~Ramamurty, G.~E. Karniadakis, and S.~Suresh,
  ``Extraction of mechanical properties of materials through deep learning from
  instrumented indentation,'' \emph{Proceedings of the National Academy of
  Sciences}, vol. 117, no.~13, pp. 7052--7062, 2020.

\bibitem{donon2020leap}
B.~Donon, B.~Donnot, I.~Guyon, Z.~Liu, A.~Marot, P.~Panciatici, and
  M.~Schoenauer, ``Leap nets for system identification and application to power
  systems,'' \emph{Neurocomputing}, 2020.

\bibitem{christie1993ieee}
R.~Christie, ``Ieee 118 bus test case,'' \emph{College of Engineering, Electric
  Engineering, University of Washington}, 1993.

\bibitem{hebbal2021multi}
A.~Hebbal, L.~Brevault, M.~Balesdent, E.-G. Talbi, and N.~Melab,
  ``Multi-fidelity modeling with different input domain definitions using deep
  gaussian processes,'' \emph{Structural and Multidisciplinary Optimization},
  vol.~63, no.~5, pp. 2267--2288, 2021.

\bibitem{neal2012bayesian}
R.~M. Neal, \emph{Bayesian learning for neural networks}.\hskip 1em plus 0.5em
  minus 0.4em\relax Springer Science \& Business Media, 2012, vol. 118.

\bibitem{gan2014optimal}
L.~Gan and S.~H. Low, ``Optimal power flow in direct current networks,''
  \emph{IEEE Transactions on Power Systems}, vol.~29, no.~6, pp. 2892--2904,
  2014.

\end{thebibliography}

\end{document}